\documentclass[journal]{IEEEtranTIE}

\usepackage{graphicx}
\usepackage{subcaption}
\usepackage{picinpar}
\usepackage{amsmath}
\usepackage{url}
\usepackage{flushend}
\usepackage[latin1]{inputenc}
\usepackage{multirow}
\usepackage[hidelinks]{hyperref}
\usepackage{enumerate}
\usepackage{siunitx}
\usepackage{tabularx, booktabs}
\usepackage{algorithm, algorithmic, amssymb}
\usepackage[backend=biber, style=ieee, sorting=none]{biblatex} 
\bibliography{Reference}

\usepackage{fancyhdr}

\renewcommand{\headrulewidth}{0pt}

\fancypagestyle{plain}{
    \fancyhf{}
    \fancyfoot[C]{\thepage}
    \renewcommand{\headrulewidth}{0pt}
}

\makeatletter
\def\ps@IEEEtitlepagestyle{
    \fancyhf{} % 清除所有页眉页脚
    \fancyfoot[C]{\thepage} % 添加页码
    \renewcommand{\headrulewidth}{0pt} % 去掉页眉横线
}
\makeatother

\begin{document}
\title{Intelligent Collaborative Optimization for Rubber Tyre Film Production Based on  Multi-path Differentiated Clipping Proximal Policy Optimization}

\author{Yinghao Ruan, Wei Pang, Shuaihao Liu, Huili Yang, Leyi Han, Xinghui Dong, \IEEEmembership{Member, IEEE}
        % <-this % stops a space
\thanks{This study was supported by the National Key Research and Development Program of China (No. 2022YFB3305300). (Corresponding author: Xinghui Dong).}
\thanks{Y. Ruan, S. Liu, and X. Dong are with the State Key Laboratory of Physical Oceanography and the Faculty of Information Science and Engineering, Ocean University of China, Qingdao, 266100. Y. Ruan is with the Faculty of Information Science and Engineering, Ocean University of China, Qingdao, 266100. W. Pang are with the School of Mathematical and Computer Sciences, Heriot-Watt University, Edinburgh, EH14 4AS. H. Yang is a Senior Engineer (Principal) at Mesnac Co., Ltd. L. Han is an Associate Senior Engineer at Mesnac Co., Ltd. (e-mail: ruan.yinghao@stu.ouc.edu.cn,  lsh7117@stu.ouc.edu.cn, xinghui.dong@ouc.edu.cn, w.pang@hw.ac.uk, yanghl@mesnac.com , hanly@mesnac.com).}
}

\maketitle
	
\begin{abstract}
The advent of smart manufacturing is addressing the limitations of traditional centralized scheduling and inflexible production line configurations in the rubber tyre industry, especially in terms of coping with dynamic production demands. Contemporary tyre manufacturing systems form complex networks of tightly coupled subsystems pronounced nonlinear interactions and emergent dynamics. This complexity renders the effective coordination of multiple subsystems, posing an essential yet formidable task. For high-dimensional, multi-objective optimization problems in this domain, we introduce a deep reinforcement learning algorithm: Multi-path Differentiated Clipping Proximal Policy Optimization (MPD-PPO). This algorithm employs a multi-branch policy architecture with differentiated gradient clipping constraints to ensure stable and efficient high-dimensional policy updates. Validated through experiments on width and thickness control in rubber tyre film production, MPD-PPO demonstrates substantial improvements in both tuning accuracy and operational efficiency. The framework successfully tackles key challenges, including high dimensionality, multi-objective trade-offs, and dynamic adaptation, thus delivering enhanced performance and production stability for real-time industrial deployment in tyre manufacturing.
\end{abstract}


\markboth{IEEE TRANSACTIONS ON INDUSTRIAL ELECTRONICS}%
{}

\definecolor{limegreen}{rgb}{0.2, 0.8, 0.2}
\definecolor{forestgreen}{rgb}{0.13, 0.55, 0.13}
\definecolor{greenhtml}{rgb}{0.0, 0.5, 0.0}

\section{Introduction}

\IEEEPARstart{U}{nder} the global trend of intelligent manufacturing transformation, the rapid development of smart manufacturing and Industry 4.0 has promoted the evolution of process industries into intelligent and networked architectures \cite{1}. Rubber tyre production, as a representative process industry, exhibits characteristics such as complex process flows, strongly coupled parameters, and high-precision quality control requirements. Specifically, tyre manufacturing involves multiple critical stages including mixing, calendering, tyre building, and vulcanization, each requiring coordinated control of multiple variables, such as temperature, pressure, and speed. Taking the tread calendering process as an example, the control precision of width and thickness directly affects the uniformity and durability of finished tyres.

The optimization of rubber tyre film production holds significant implications for enhancing product quality, reducing production costs, and advancing green manufacturing. Firstly, precise dimensional control of the film is fundamental to ensuring structural integrity and driving safety of tyres, directly influencing their dynamic balance and wear performance. Secondly, coordinated parameter optimization during production can significantly reduce material waste and energy consumption, aligning with sustainable development goals. Furthermore, the introduction of intelligent optimization methods facilitates the development of adaptive and scalable production systems, promoting the transition of the tyre industry toward digitalized and flexible manufacturing paradigms. This provides a technical framework for the intelligent transformation of process industries.

\begin{figure}[h]
\centerline{\includegraphics[width=0.98\columnwidth]{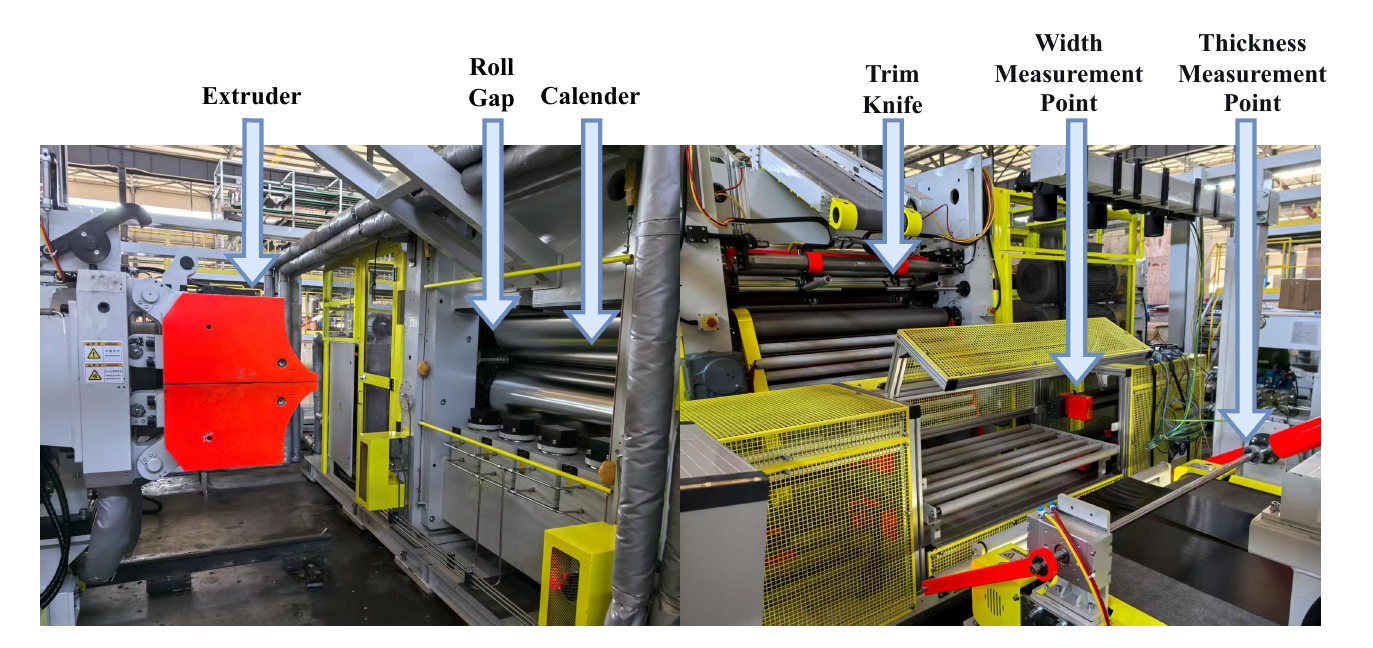}}
\caption{Rubber Tyre Film Manufacturing Equipment.}
\label{fig1}
\end{figure}

Traditional control methods face challenges in achieving dynamic coordination and global optimization across industrial processes.  In complex multi-stage production systems, there are implementations that leverage the integration of IoT \cite{2}, digital twins \cite{3}, multi-agent systems (MAS) \cite{4}, and deep reinforcement learning (DRL) \cite{5}, demonstrating potential in enhancing overall synergistic efficiency and achieving dynamic optimization.

Digital twins facilitate real-time mapping of physical systems, supporting parameter optimization and anomaly diagnosis, while DRL techniques, including Proximal Policy Optimization (PPO) \cite{6} and Twin Delayed Deep Deterministic Policy Gradient (TD3) \cite{7}, have shown strong performance in continuous control and collaborative optimization. Additionally, Transformer and graph neural networks (GNNs) enhance high-dimensional state representation and cross-domain generalization. Nevertheless, tyre production remains a challenging domain due to its high dimensionality, strong coupling, and nonlinear dynamics, demanding more robust and adaptive collaborative algorithms.

Proximal Policy Optimization (PPO), while valued for its stability in industrial control, exhibits limitations in high-dimensional state-action spaces which is typical in tyre production. Its fixed clipping mechanism proves inadequate under dynamic operating conditions, leading to optimization instability, while difficulties in balancing exploration-exploitation and sensitivity to reward design further constrain its effectiveness in collaborative optimization tasks.

To overcome these challenges, this study introduces a multi-path differentiated clipping mechanism that enables dimension-aware policy updates. The enhanced algorithm demonstrates accelerated convergence, improved stability in dynamic environments, enhanced multi-objective optimization capability, and greater robustness, thus offering a more reliable control solution for rubber tyre manufacturing.

The main contributions of this work are listed as follows: 

\begin{itemize}

\item We propose an autonomous optimization framework integrating deep learning and reinforcement learning to achieve dynamic perception, quality prediction, and self-adaptive parameter control in industrial processes.

\item We introduce a predict model that effectively captures both short-term local dependencies and long-term trends in multi-dimensional process parameters.

\item We develop a Multi-path Differentiated Clipping Proximal Policy Optimization (MPD-PPO) algorithm, employing a multi-path policy network and differentiated clipping mechanism to enhance training stability and optimization efficiency in high-dimensional action spaces.

\item We design a composite reward function for rubber tyre film production, incorporating exponential decay and hyperbolic tangent components to balance quality accuracy, operational stability, and energy efficiency in multi-objective optimization.
\end{itemize} 

The remainder of this paper is structured as follows. Section \uppercase\expandafter{\romannumeral2} reviews related work in industrial reinforcement learning and predictive modeling. Section \uppercase\expandafter{\romannumeral3} introduces the proposed methodology, including the LSTNet predictive model and the MPD-PPO algorithm with its multi-path architecture. Section \uppercase\expandafter{\romannumeral4} presents experimental results and analysis on rubber tyre film production. Finally, Section \uppercase\expandafter{\romannumeral5} concludes the paper and discusses future directions.

\section{Related Work}

\subsection{Deep Learning for Industrial Prediction }
Deep learning has revolutionized industrial predictive capabilities by overcoming the limitations of traditional feature engineering. Through successive architectural innovations, it has addressed diverse challenges in industrial data processing.

In industrial image processing, Convolutional Neural Networks (CNNs) \cite{8} are widely applied in tasks such as surface defect detection and target positioning. Evolving from LeNet \cite{9} to AlexNet \cite{10}, ResNet \cite{11}, and DenseNet \cite{12}, these architectures have significantly advanced feature extraction for visual inspection. Lightweight networks like MobileNet \cite{13} and ShuffleNet \cite{14} enable real-time edge applications, though with accuracy-efficiency trade-offs. Recently, Vision Transformers (ViT) \cite{15} have demonstrated remarkable capabilities in capturing long-range dependencies, despite higher computational costs.

For temporal data processing, Recurrent Neural Networks (RNN) \cite{16} and their variants dominate, with Long Short-Term Memory (LSTM) \cite{17} and Gated Recurrent Units (GRU) \cite{18} effectively capturing long-term dependencies, while Temporal Convolutional Networks (TCN) \cite{19} and Transformers \cite{20} offer advantages in parallel processing and complex pattern capture respectively, and graph neural networks (GNNs) \cite{21} enhance high-dimensional state representation and cross domain generalization.

Despite these advancements, deep learning models primarily function as passive analyzers in industrial settings. While excelling at feature extraction and pattern recognition, they lack inherent capabilities for autonomous decision-making and action execution in dynamic environments.

\subsection{Reinforcement Learning for Industrial Control }

Reinforcement Learning (RL) has become a prominent framework for sequential decision-making in complex industrial environments, effectively tackling control and optimization problems beyond the reach of traditional methods.

The RL landscape includes several algorithm families tailored to industrial applications. Value-based methods such as Deep Q-Networks (DQN) \cite{22} and its variants (Double DQN \cite{23}, Dueling DQN \cite{24}) excel in discrete control tasks like production scheduling, yet are less suitable for continuous action spaces common in industrial systems.

Policy-based methods overcome this by directly optimizing parameterized policies. While REINFORCE offers a foundational approach, advanced algorithms like Trust Region Policy Optimization (TRPO) \cite{25} and Proximal Policy Optimization (PPO) \cite{6} ensure training stability, which is essential for safety-sensitive industrial settings. PPO, in particular, is widely adopted due to its clipped objective and robust performance.

Actor-critic methods merge value and policy-based advantages, proving effective for complex industrial tasks. Deep Deterministic Policy Gradient (DDPG) \cite{26} extends DQN to continuous domains, enabling applications in robotic and process control. Twin Delayed DDPG (TD3) \cite{7} enhances robustness via clipped double Q-learning and delayed updates, while Soft Actor-Critic (SAC) \cite{27} promotes exploration through entropy maximization.

Multi-Agent RL (MARL) \cite{28} methods address collaborative industrial systems. Algorithms such as VDN \cite{29} and QMIX \cite{30} solve credit assignment in multi-agent settings, facilitating coordination in applications like automated logistics and smart energy grids.

Model-based RL (MBRL) \cite{31} and Offline RL \cite{32} offer improved sample efficiency through internal simulation and effective utilization of historical datasets, supporting multi-scale decision-making and reducing retraining needs.

Despite the above mentioned advances, key challenges remain. RL algorithms are often sample-inefficient, requiring large amounts of costly or risky operational data. Reward design lacks systematic methodology, and safety during exploration remains a critical barrier-especially in real-world industrial deployments.

\section{A Predictive-Optimization Closed Loop Method for Industrial Control}

Modern industrial processes like tyre manufacturing exhibit strongly-coupled parameters across stages, making multi-objective optimization essential yet challenging. This paper proposes an integrated LSTNet and MPD-PPO framework to overcome suboptimal sequential methods. LSTNet captures temporal dependencies for quality prediction, while MPD-PPO's multi-branch policy enables parallel objective optimization, achieving adaptive decision-making under complex constraints control precision and system efficiency in dynamic industrial environments.

\subsection{Overview}

\begin{figure*}[h]
\centerline{\includegraphics[width=2.0\columnwidth]{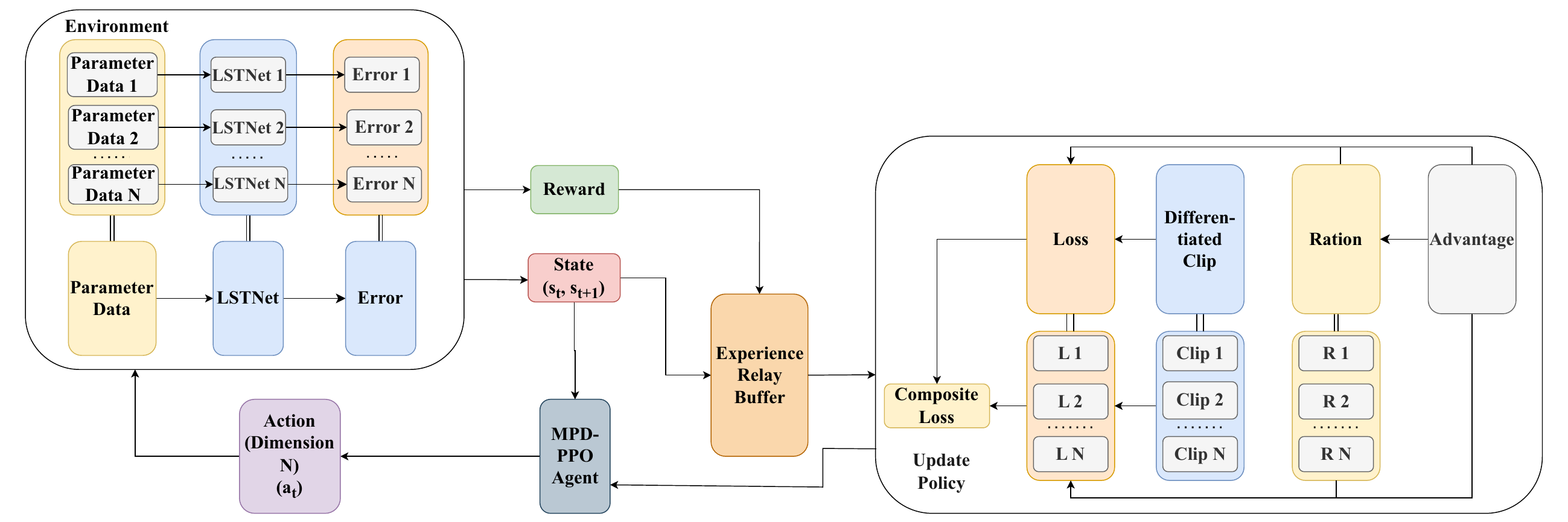}}
\caption{The Workflow Diagram Of The Predictive-Optimization Closed Loop Method. }
\label{fig1}
\end{figure*}

The  workflow of the collaborative optimization method integrating LSTNet and MPD-PPO for industrial control  is illustrated in Fig. 2.  The initialization phase involves constructing an MPD-PPO agent, which consists of an actor network with \emph{N} independent pathways and a shared critic network. Each action pathway's architecture is individually configurable and tailored to the physical properties of its specific actuator. For continuous action spaces, a trainable standard deviation parameter is assigned to each pathway, with its initial value determined by the actuator's precision requirements. Before training commences, an experience replay buffer is initialized using a partitioned storage strategy, which maintains separate historical action probability distributions for every action pathway.

At each training timestep $t$, the environmental state $s_t$ is fed into the policy network. Each action pathway then operates in parallel: shared network layers first extract foundational features, which are then processed by pathway-specific fully connected layers to produce the action mean values. Subsequently, actions $a_t$ are stochastically sampled using the current standard deviation.

The outputs from all pathways are concatenated to form the final action vector $a_t$. The interaction with the environment then proceeds as follows: firstly, the environment decodes the continuous action vector $a_t$ and reconstructs the input features for the downstream forecasting module. This is achieved by mapping action values to incremental set-points of process parameters, which accordingly alters the original data to generate a new time-series sample. This sample is then fed into the LSTNet network to acquire the next-step estimates of parameters. Finally, the prediction error is normalized, and this normalized error serves as an instantaneous reward signal and is also used to form the subsequent state vector $s_{t+1}$.

The resulting transition tuple is stored in the experience replay buffer. The policy update phase is triggered as soon as the number of accumulated trajectories exceeds a predefined threshold. This "interaction-prediction-reward-storage-update" loop iterates continuously until the policy converges to an equilibrium state, thereby achieving high-precision cooperative optimization of multiple process parameters.

\subsection{A Deep Learning Framework for Industrial Prediction  }

Deep learning models offer strong predictive power and latent pattern identification in complex data. For time series prediction, architectures such as CNNs, LSTMs, and self-attention mechanisms are commonly chosen for their ability to capture temporal dynamics and long-range dependencies, with selection considering adaptability, learning efficiency, and computational complexity.

\subsubsection{Architecture for LSTNet}
As shown in Fig. 2, the proposed LSTNet integrates CNNs, LSTM, and a Skip-GRU module to concurrently capture multi-scale temporal patterns in industrial data. Utilizing a sliding window approach, it processes input sequences for localized feature extraction while preserving long-term temporal dependencies.

\subsubsection{Hierarchical Feature Extraction Pipeline}
The network employs a hierarchical feature extraction process, beginning with stacked 1D convolutional and pooling layers to capture multi-scale local patterns. The resulting features are then modeled for long-term dependencies by a unidirectional LSTM. Layer normalization and dropout are integrated throughout to ensure stable training and prevent overfitting. 

\subsubsection{Skip-GRU Module for Multi-Scale Temporal Modeling}
In tyre manufacturing, a novel model integrates an LSTM pathway with a dedicated Skip-GRU module. The Skip-GRU captures multi-scale periodic patterns via skip-connected windows, while the LSTM models sequential dynamics. Their combined outputs are fused to comprehensively represent temporal dependencies, enabling accurate production prediction.

In summary, LSTNet effectively combines convolutional feature extraction, recurrent temporal modeling, and multi-scale skip connections to handle complex time series data, capturing both short-term fluctuations and long-term trends.

\begin{figure}[h]
\centerline{\includegraphics[width=0.98\columnwidth]{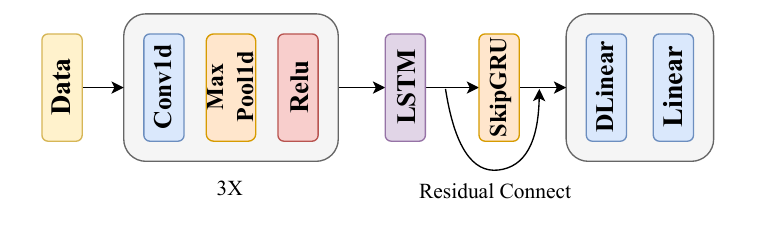}}
\caption{Architecture of the LSTNet Model. }
\label{fig1}
\end{figure}

\subsection{Multi-path Differentiated Clipping Proximal Policy Optimization}

This study proposes an enhanced reinforcement learning algorithm named Multi-path Differentiated Clipping Proximal Policy Optimization (MPD-PPO), tailored for collaborative optimization in industrial processes. Building upon the Proximal Policy Optimization (PPO) framework, MPD-PPO introduces a multi-branch policy network architecture to address the limitations of uniform policy updates in traditional PPO when applied to industrial control.  
Each actuator corresponds to an independent policy branch, with the action distribution for the \textit{ i-th} actuator defined as follows: 
\begin{equation}\pi_{\theta_i}(a_i,s)=N(\mu_i(s),\sigma^2_i)\label{eq}.\end{equation}
In the policy representation, the meaning of each variable is as follows:
$\pi_{\theta_i}(a_i,s)$: probability distribution of action  $a_i$ given state $s$  for the \textit{i-th}  actuator;  $\theta_i$: trainable parameters of the policy network corresponding to the \textit{i-th} actuator; $a_i$: continuous action output of the \textit{i-th}  actuator; $s$: global environmental state; $\mu_i(s)$: state-dependent mean action generated by the neural network; $\sigma^2_i$: variance controlling the exploration magnitude.

A key innovation of  MPD-PPO is the differentiated Clip mechanism, which adaptively constrains policy updates per actuator based on dynamic response characteristics. The objective function for each path is formulated as follows: 
\begin{equation}L_i^{CLIP}=E_t[min(r^i_t(\theta_i)A_t,clip(1-\epsilon_i,1+\epsilon_i)A_t)],
\end{equation}
where the importance sampling ratio $r^i_t(\theta_i)=\frac{\pi_{\theta_i}(a_t^i,s_t^i)}{\pi_{\theta_i,old}(a_t^i,s_t^i)}$ and the clipping threshold $\epsilon_i$ are configured dynamically. 

The advantage function calculation employs pathway-specific discounted return estimation. For the \textit{i-th} action pathway, the advantage estimate is given as follows:
\begin{equation}
    A^t_i=G^t_i-V_i(s_t),
\end{equation}
where the value function $V_i$ is generated by the shared Critic network, $G^t_i$ employs generalized advantage estimator  (GAE) \cite{33} for computation.  This design enables different action pathways to adopt distinct discount factors  $\gamma_i$ according to their response characteristics.

Policy optimization begins by calculating and standardizing the Monte Carlo discounted returns. Following this, these returns are combined with the state-values from the current policy to compute advantage estimates, which drive the gradient update.

Within the updating loop, empirical data is reshuffled and partitioned into mini-batches. For each batch, a forward pass is conducted to compute the importance ratios for each action branch. To respect the distinct dynamics of individual process parameters, branch-specific clipping ranges are applied, and the resulting surrogate losses are weighted according to real-time control priorities (specified by weights $w_i$). The total loss function $L_t$ combines the weighted surrogate loss, the value function loss (with coefficient $c_1$), and an entropy regularization term (with coefficient $c_2$). This loss is optimized via backpropagation with gradient accumulation, after which the policy parameters are updated, and the old policy is synchronized for the subsequent data collection phase. The update procedure of MPD-PPO is presented in Algorithm 1.

\begin{algorithm}
\caption{MPD-PPO Update Policy}
\begin{algorithmic}[1]
\REQUIRE Learning rate $\alpha$, discount factor $\gamma$, epochs $K$, clip ranges $\epsilon_i$, batch size $B$
\ENSURE Optimized policy networks $\{\pi_{\theta_i}\}_{i=1}^N$

\STATE Initialize policy networks $\{\pi_{\theta_i}\}_{i=1}^N$, value network $V_{\phi}$, optimizer
\STATE Initialize experience replay buffer $\mathcal{B}$

\FOR{each training episode}
    \STATE Collect trajectory $\tau = \{(s_t, a_t, r_t, s_{t+1})\}$ and store in $\mathcal{B}$
    
    \IF{$\mathcal{B}$ reaches predefined size}
        \STATE Compute discounted returns: $R_t = \sum_{k=0}^{\infty} \gamma^k r_{t+k}$
        \STATE Compute advantages: $A_t = R_t - V_{\phi}(s_t)$
        
        \FOR{$k = 1$ \TO $K$}
            \STATE Sample random batch: $\mathcal{B}_{batch} \sim \mathcal{B}$
            
            \FOR{each policy network $\pi_{\theta_i}$}
                \STATE Compute probability ratio: $\rho_i(\theta) = \frac{\pi_{\theta_i}(a_t|s_t)}{\pi_{\theta_{i, old}}(a_t|s_t)}$
                \STATE Compute clipped loss: 
                $L_i^{CLIP} = \mathbb{E}[\min(\rho_i(\theta)A_t, \text{clip}(\rho_i(\theta), 1-\epsilon_i, 1+\epsilon_i)A_t)]$
            \ENDFOR
            
            \STATE Compute value loss: $L^{VF} = \mathbb{E}[(V_{\phi}(s_t) - R_t)^2]$
            \STATE Compute entropy bonus: $L^{ENT} = \mathbb{E}[H(\pi_{\theta}(s_t))]$
            \STATE Total loss: $L^{TOTAL} = \sum_{i=1}^N w_i L_i^{CLIP} + c_1 L^{VF} + c_2 L^{ENT}$
            \STATE Gradient update: $\theta \leftarrow \theta - \alpha \nabla_{\theta} L^{TOTAL}$
        \ENDFOR
        
        \STATE Clear buffer $\mathcal{B}$
    \ENDIF
\ENDFOR
\end{algorithmic}
\end{algorithm}

\subsection{Reward Function }

This study employs a composite reward function where individually computed objective-specific rewards are normalized and aggregated. This ensures balanced optimization across objectives with divergent numerical scales, promoting stable training.

\subsubsection{Error Reward} an exponential decay function is employed to increase sensitivity to small errors and promote continuous precision improvement:
\begin{equation}
R_e = 2.0 \cdot \exp(-e_t).
\end{equation}
\subsubsection{Progress Reward} To maintain gradient stability and ensure the agent receives sustained incentives while enabling smooth and stable policy optimization, the hyperbolic tangent function constrains the reward values within the interval $(-0.3, 0.3)$, thereby motivating the agent to surpass historical performance benchmarks:
\begin{equation}
    R_p = 0.3 \cdot \tanh(e_{\text{best}} - e_t).
\end{equation}
\subsubsection{Action Penalty} this term penalizes large adjustments in process parameters to encourage smooth control operations:
\begin{equation}
    P_a = -0.05 \cdot (a_t - a_{t-1})^2.
\end{equation}
\subsubsection{Steady-State Reward} this term provides incremental positive feedback when the system operates near the target value, thereby enhancing operational stability:
\begin{equation}
    R_s = 0.5 \cdot (\tau - e_t).
\end{equation}

The variables used in the above equations are defined as follows:
\begin{itemize}
\item $e_t$: the absolute tracking error at time step $t$, defined as $|y_t - y_{\text{target}}|$, where $y_t$ is the current predicted value (e.g., width or thickness) and $y_{\text{target}}$ is the desired value.
\item $e_{\text{best}}$: the minimum observed tracking error throughout the current training episode.
\item $a_t$: the current value of the adjusted process parameter (e.g., blade gap or roller distance).
\item $a_{t-1}$: the previous value of the process parameter, used to compute the control action magnitude.
\item $\tau$: a constant threshold that defines the tolerance range for the steady-state region.
\end{itemize}

The overall reward at each time step is computed as the sum of these components:
\begin{equation}
R_t = R_e + R_p + P_a + R_s.
\end{equation}

The reward coefficients were calibrated through sensitivity analysis to ensure balanced collaborative optimization, with total rewards clipped to a predefined range for numerical stability during training. This structured reward design effectively integrates tracking precision, behavioral smoothness, and steady-state performance, addressing critical requirements of industrial control systems.

\subsection{Simulation Environment for Industrial Collaborative Optimization  }

This study develops a high-fidelity simulation environment that serves as the core interface for reinforcement learning agents to interact with a simulated industrial production system. Tailored for collaborative optimization, the environment features a continuous, multi-dimensional action space for fine-grained parameter control and a state representation that captures temporal dependencies of key process variables. A composite reward function enables the balance of competing objectives, with dynamically adjustable weights to prioritize goals in real-time. Episode termination is governed by achieving predefined precision thresholds across all target metrics, ensuring comprehensive optimization. Built on real production data with a modular architecture, the environment provides a robust testbed for developing and transferring effective control policies to real-world applications.

\section{Performance Validation for Rubber Tyre Film Production}

Based on the integrated deep learning and reinforcement learning framework, this section presents empirical validation in the rubber tyre film calendering process. This scenario exhibits strongly coupled and nonlinear control of film width and thickness - critical quality parameters that challenge conventional methods.

The study innovatively combines temporal prediction models with the MPD-PPO algorithm: the former captures complex parameter dependencies to predict quality indicators, while the latter dynamically adjusts control strategies for optimal parameters under multiple constraints. Comparative experiments targeting width (480$mm$$/$380$mm$) and thickness (3.0$mm$$/$2.2$mm$) objectives demonstrate MPD-PPO's convergence behavior and optimization efficacy. Results show robust performance across all targets, providing empirical support for industrial applications.

 \subsection{Data Acquisition }

To thoroughly investigate the relationship between rubber film width and various machine/environmental parameters, real-world process data were collected from the inner-liner production line. MESNAC\footnote{\url{https://en.mesnac.com/}}, a key strategic partner and major supplier of intelligent manufacturing systems to the global tyre industry, provided continuous \emph{in-situ} measurements via industrial IoT sensors and monitoring devices. The company's strong R\&D capabilities ensured the reliability and precision of the dataset used in this study. As illustrated in Fig. 1, the manufacturing process begins with compound plasticization in an extruder. The material is then discharged onto a conveyor and fed into a calender to form a continuous sheet. This sheet is subsequently cropped to the required width by edge-trimming knives, drawn through pull rolls, and finally delivered to the take-off conveyor. Thermal expansion and contraction cause the width and thickness of the trimmed and drawn strip to fluctuate until the sheet stabilizes at the take-off section. A critical constraint arises from the equipment setup: width and thickness gauges are installed several meters downstream on the take-off belt, introducing a measurement delay. This delay not only prevents real-time monitoring of the current process state but also leads to material waste during product-specification changes. Consequently, the continuous process data collected from this line (Fig. 3) provide a valuable foundation for developing intelligent predictive models. According to plant engineers, film width and thickness are governed by the machine parameters listed in Tables \uppercase\expandafter{\romannumeral1} and \uppercase\expandafter{\romannumeral2}. By tuning these parameters, it is possible to achieve concurrent optimization of both width and thickness.

\begin{table}[htbp]
\centering
\caption{Key Factors Influencing the Width of Rubber Tyre  Film.}
\label{tab:three_line}
\begin{tabularx}{0.48\textwidth}{>{\raggedright\arraybackslash}X c}
\toprule
\textbf{Technical Parameter} & \textbf{Function} \\
\midrule
Main-motor current of calender ($A$) & / \\
Calender line-speed ($mm$) & / \\
Discharge temperature of calender ($^\circ$C) & / \\
Draw-roll speed of calender ($mm$) & / \\
Conveyor-belt speed ($mm$) & / \\
Actual knife-to-knife distance of edge-trimming unit ($mm$) & Control variable \\
Top-roll temperature of calender ($^\circ$C) & / \\
Bottom-roll temperature of calender  ($^\circ$C) & / \\
Main-motor current of extruder ($A$) & / \\
Extruder screw speed ($mm$) & / \\
Extruder head pressure ($N$) & / \\
Extruder head temperature ($^\circ$C) & / \\
Extruder screw-tip pressure ($N$) & / \\
Extruder head temperature-control set-point ($^\circ$C) & / \\
Extruder plasticizing zone-1 temperature ($^\circ$C) & / \\
Extruder plasticizing zone-2 temperature ($^\circ$C)& / \\
Extruder barrel temperature ($^\circ$C) & / \\
Extruder screw temperature ($^\circ$C) & / \\
\bottomrule
\end{tabularx}
\end{table}

\begin{table}[htbp]
\centering
\caption{Key Factors Influencing the Thickness of Rubber Tyre Film. }
\label{tab:three_line}
\begin{tabularx}{0.48\textwidth}{>{\raggedright\arraybackslash}X c}
\toprule
\textbf{Technical Parameter} & \textbf{Function} \\
\midrule
Extruder screw speed ($mm$) & / \\
Extruder main-motor current ($A$)& / \\
Extruder outlet temperature ($^\circ$C) & / \\
Extruder outlet pressure ($N$)& / \\
Extruder screw-tip pressure ($N$)& / \\
Screw temperature-control set-point ($^\circ$C) & / \\
Plasticizing zone-1 temperature ($^\circ$C) & / \\
Plasticizing zone-2 temperature ($^\circ$C) & / \\
Extruder head temperature ($^\circ$C) & /\\
Calender line-speed ($mm$) & / \\
Calender main-motor current ($A$)& / \\
Sheet temperature after calender ($^\circ$C) & / \\
DS (drive-side) roll gape ($mm$)& Control variable \\
OS (operator-side) roll gap ($mm$)& Control variable \\
Edge-trimming knife spacing ($mm$) & / \\
Prick-roll temperature ($^\circ$C) &  /\\
Extrusion-section temperature ($^\circ$C) & / \\
\bottomrule
\end{tabularx}
\end{table} 

\subsection{Experimental Results Of Prediction}

 For width and thickness prediction, the LSTNet was trained using the Adam optimizer with Mean Absolute Error (MAE) as the loss function. We trained two separate prediction models on the collected width and thickness datasets, respectively. After 100 training epochs with a learning rate of 0.001 and a batch size of 1024, both models achieved convergence. Subsequent evaluation on the test set yielded the average prediction error and qualification rate, which are summarized in Table \uppercase\expandafter{\romannumeral3}. The qualification rate is defined as the proportion of samples whose prediction errors for both width and thickness are within the tolerance ranges specified by the factory ($\pm$1 mm for width, $\pm$0.05 mm for thickness).
\begin{table}[htbp]
\centering
\caption{Predicted Results Of LSTNet for Width and Thickness.}
\label{tab:three_line}
\begin{tabularx}{0.48\textwidth}{>{\raggedright\arraybackslash}X c c c}
\toprule
\textbf{Data} & \textbf{MAE  ($mm$)}   & \textbf{RMSE  ($mm$)}  & \textbf{Qualification Rate}  \\
\midrule
Width & 1.865$\pm$ 0.099 & 2.458$\pm$0.0.138 & 0.452$\pm$0.073\\
Thickness  & 0.204$\pm$0.053 &  0.316$\pm$0.074 &  0.579$\pm$0.024 \\
\bottomrule
\end{tabularx}
\end{table} 

To systematically assess the effectiveness and generalization of the proposed prediction model, this study conducts an empirical comparative analysis using tyre film width prediction as a representative production task. All models were trained under identical hyperparameters, including a learning rate of 0.001 and a batch size of 1, 024, and for 100 epochs per run, with results averaged over 10 independent trials. Four models were used for comparison, including  BILSTM\cite{34}, Linear Regression (LR), TCN\cite{19} and Informer\cite{35}. The models were evaluated on four metrics:  MAE, RMSE, and Qualification Rate (QR). The comparative experimental results are summarized in Table \uppercase\expandafter{\romannumeral4}. 
\begin{table}[htbp]
\centering
\caption{ A Comparative Study of Predictive Performance Between LSTNet and Other Networks. }
\label{tab:three_line}
\begin{tabularx}{0.48\textwidth}{>{\raggedright\arraybackslash}X c c c}
\toprule
\textbf{Net} & \textbf{MAE}  ($mm$)   & \textbf{RMSE}  ($mm$)  & \textbf{QR}  \\
\midrule
BILSTM \cite{34} & 19.125$\pm$0.654 & 26.955$\pm$1.006  & 0.097$\pm$0.052 \\
LR  &  3.782$\pm$1.212  &  4.604$\pm$1.449  & 0.206$\pm$0.068 \\
TCN \cite{19} & 10.188$\pm$0.759 & 14.071$\pm$1.094 & 0.129$\pm$0.055  \\
Informer \cite{36} & 13.689$\pm$0.872 & 19.551$\pm$1.257 &  0.095$\pm$0.018 \\ 
\textbf{LSTNet} & \textbf{1.865$\pm$ 0.099} & \textbf{ 2.458$\pm$0.0.138}  & \textbf{0.452$\pm$0.073}  \\
\bottomrule
\end{tabularx}
\end{table} 

\subsection{Experimental Results Of Optimization }

The MPD-PPO algorithm utilizes separate policy networks for width and thickness adjustment, complemented by an adaptive clipping mechanism. This hierarchical architecture features shared layers for modeling global interactions and dedicated branches for process-specific optimization.

In the clipping mechanism design, MPD-PPO dynamically sets clipping thresholds based on parameter sensitivity analysis:

\begin{itemize}

\item Width control (larger-scale adjustments): we use a relaxed clipping threshold range of (0.8, 1.2)

\item  Thickness control (finer-scale adjustments): we employ a stricter clipping threshold range of (0.9, 1.1)

\end{itemize}

For width control, a looser clipping bound allows for larger-magnitude adjustments, whereas for thickness, a tighter bound prevents aggressive updates to ensure stable fine-tuning. This study investigates the impact of training steps per episode on the optimization performance of the MPD-PPO agent. We systematically compared configurations of 50 and 100 training steps per episode to evaluate their effects on convergence behavior and solution quality, where at each time step the agent selects an action.  To enhance the robustness of the findings, we conducted five independent experimental trials and consolidated the resulting optimization curves into an intuitive interval representation for visualization. The experimental outcomes after 50 and 100 training episodes are presented in Fig. 4, Fig. 5, and Table \uppercase\expandafter{\romannumeral5}. It is noteworthy that the thickness curve appears wider post-convergence in the figure. This is a visual effect resulting from the high resolution of the y-axis scale, which is necessitated by its small absolute numerical range, rather than an indication of increased actual volatility.

\begin{table}[htbp]
\centering
\caption{Performance of the MPD-PPO Algorithm.}
\label{tab:three_line}
\begin{tabularx}{0.48\textwidth}{>{\raggedright\arraybackslash}X c c}
\toprule
\textbf{Train Step} & \textbf{Target (Width, Thickness)}   ($mm$) & \textbf{Average Optimize Step} \\
\midrule
\multirow{4}*{50} & (480, 3.0) & 23 \\
                  & (480, 2.2) & 18 \\
                  & (380, 3.0) & 14 \\
                  & (380, 2.2) & 19 \\
\midrule
\multirow{4}*{100} & (480, 3.0) & 22 \\
                   & (480, 2.2) & 19 \\
                   & (380, 3.0) & 14 \\
                   & (380, 2.2) & 16 \\
\bottomrule
\end{tabularx}
\end{table}

 \begin{figure}
    \centering
    \includegraphics[width=0.98\linewidth]{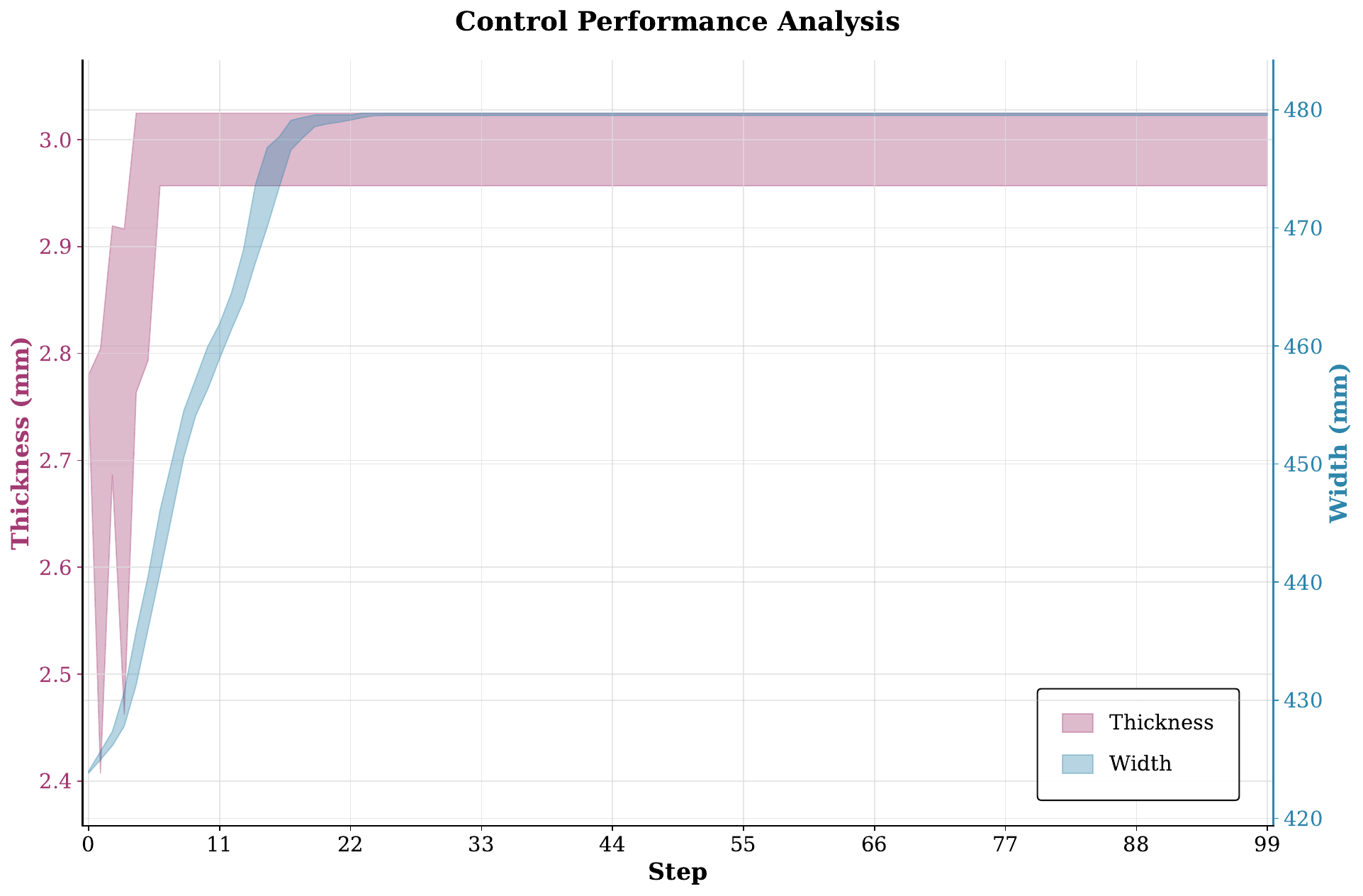}
    \caption{MPD-PPO Optimization Performance at 50 Training Steps. }
    \label{fig:placeholder}
\end{figure}
  \begin{figure}
    \centering
    \includegraphics[width=0.98\linewidth]{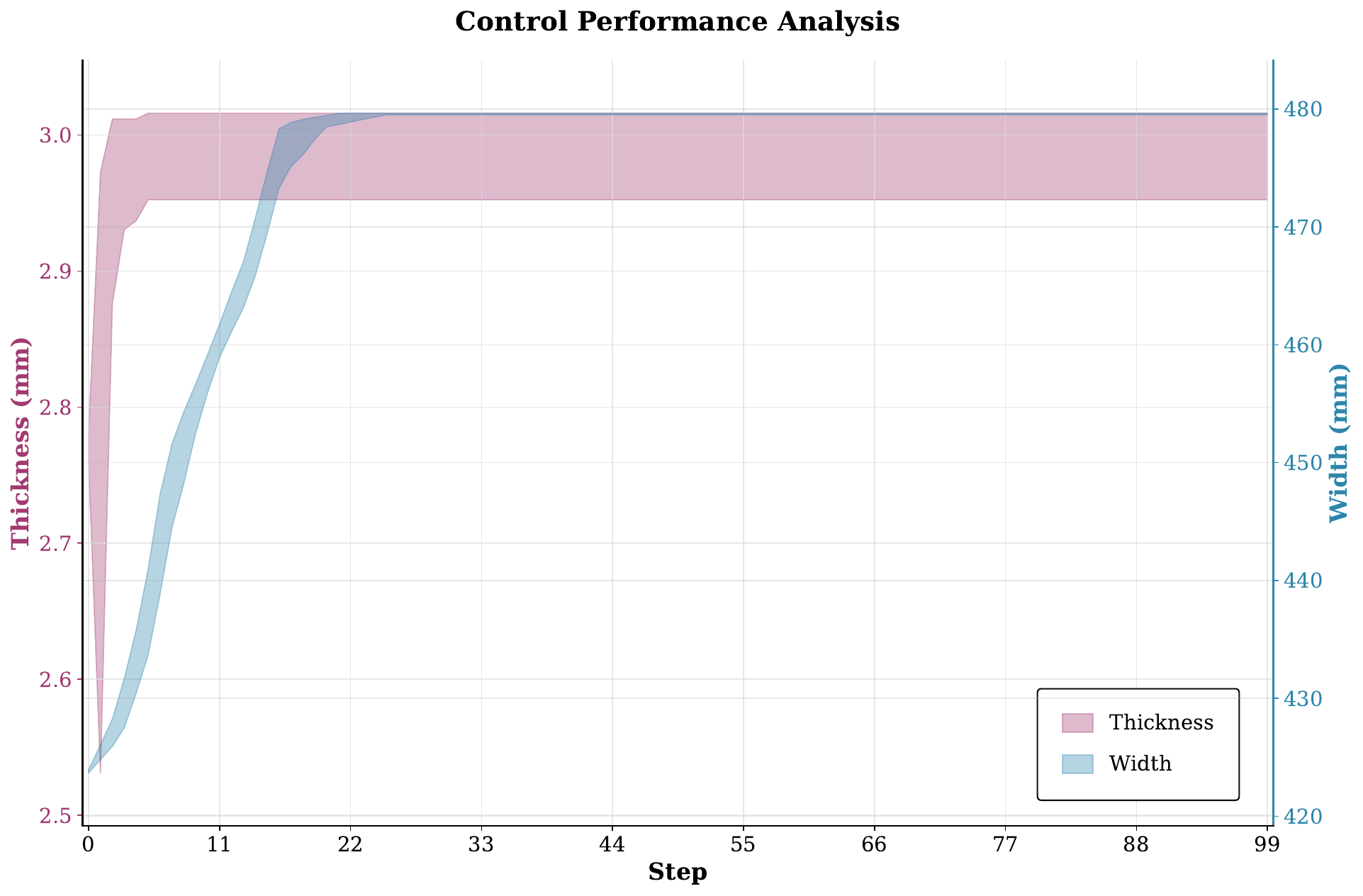}
    \caption{MPD-PPO Optimization Performance at 100 Training Steps. }
    \label{fig:placeholder}
\end{figure}
At 50 training steps, the optimization results demonstrate considerable effectiveness, though minor oscillations are observed in localized regions. By 100 training steps, the agent exhibits significantly improved performance, consistently achieving optimization targets within approximately 20 steps  across all evaluated episodes. These results indicate robust convergence and high sample efficiency of the proposed method.

\subsection{Ablation Study }

This section evaluates the effectiveness of the multi-branch network architecture and the proposed differential clipping strategy against two baseline methods: PPO and MPD-PPO. During agent training, a consistent protocol was used, with each agent trained for 100 episodes, each comprising 100 time steps.

\subsubsection{The Influence Of Multi-Branch Network}
In this experiment, we replaced the original single-network architecture of the PPO algorithm with a two-branch network, comprising separate width and thickness action networks to investigate the efficacy of the branched architecture. Representative results for the typical operating condition of 480 mm width and 3.0 mm thickness are presented for comparative analysis. Table \uppercase\expandafter{\romannumeral6} summarizes the quantitative results. Figures 6 and 7 present the performance of PPO with Multi-Branch Network and standard PPO, respectively, following the MPD-PPO results in Fig 5.

Based on the experimental results, replacing the original policy network with a dual-branch architecture yields significant performance improvements. Although the width optimization still does not reach the target value, its deviation is substantially reduced compared to conventional PPO. Furthermore, both width and thickness optimization trajectories demonstrate smoother profiles relative to the traditional approach.
\begin{table}[htbp]
\centering
\caption{Optimization Results Of Removing Differentiated Clip. }
\label{tab:three_line}
\begin{tabularx}{0.48\textwidth}{>{\raggedright\arraybackslash}X c}
\toprule
\textbf{ Algorithm}  & \textbf{Average Optimize Step} \\
\midrule
MPD-PPO  & 22 \\
PPO With Multi-Branch Network  & 100 (Failed)\\
PPO  & 100 (Failed) \\
\bottomrule
\end{tabularx}
\end{table} 
  \begin{figure}[h]
    \centering
    \includegraphics[width=0.98\linewidth]{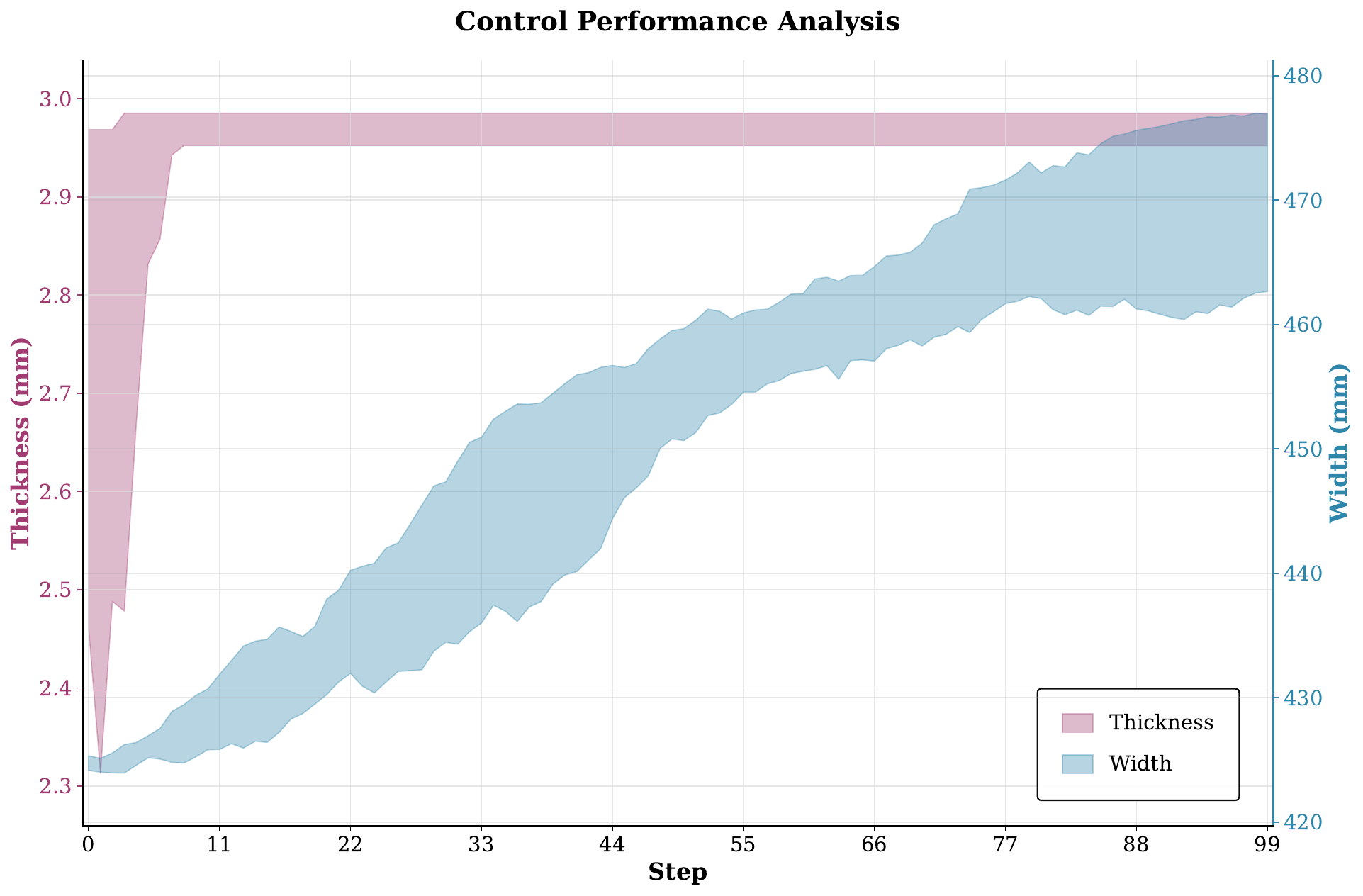}
    \caption{Optimization Results of PPO with Multi-Branch Network. }
    \label{fig:placeholder}
\end{figure}
\begin{figure}[h]
    \centering
    \includegraphics[width=0.98\linewidth]{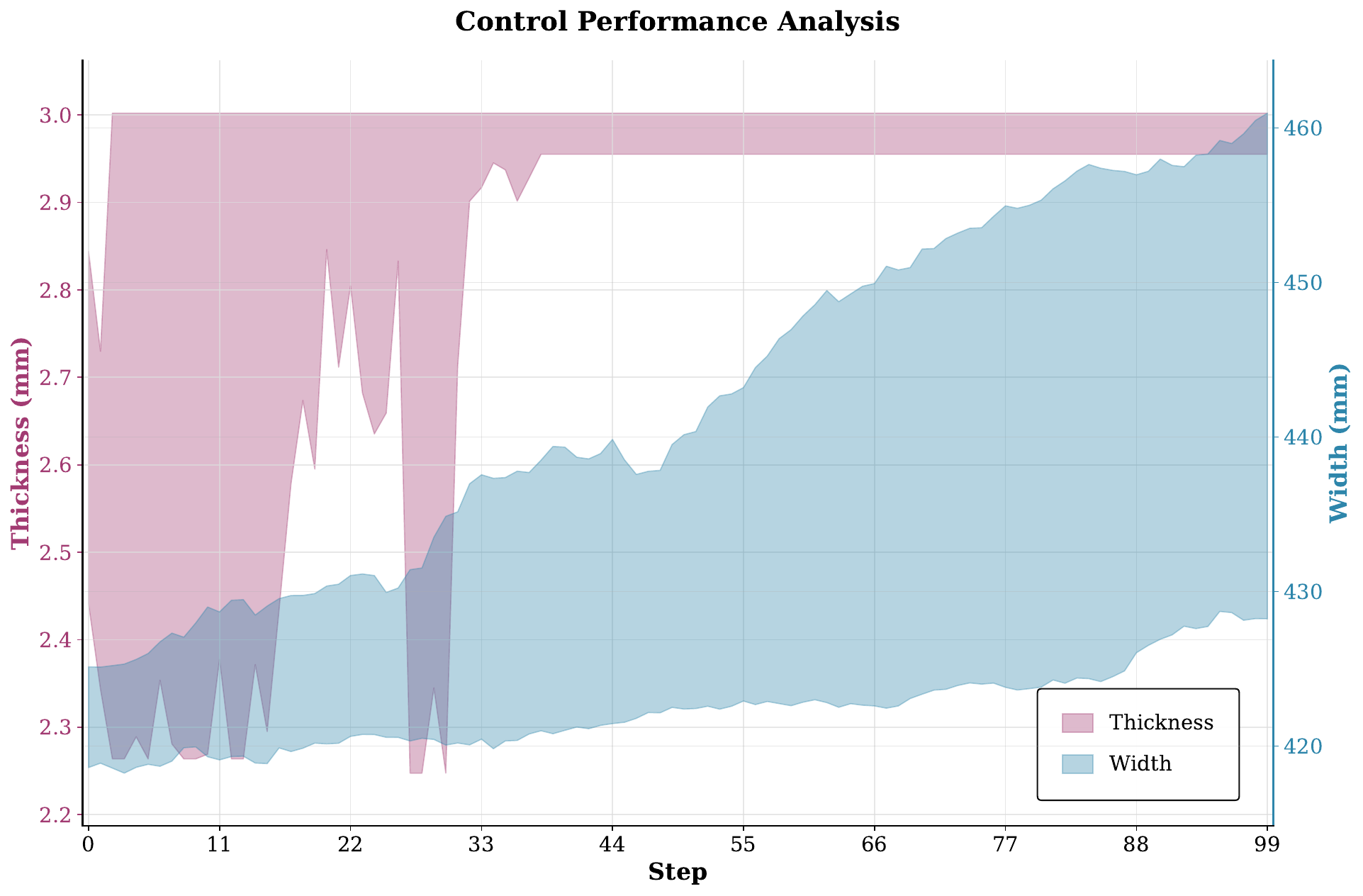}
    \caption{Optimization Results of  PPO Algorithm. }
    \label{fig:placeholder}
\end{figure}

\subsubsection{The influence of Differentiated Clipping}
This experiment sets identical clipping thresholds for both branches to investigate the efficacy of the differentiated clipping strategy. Representative results for a width of 480 mm and a thickness of 3.0 mm are presented for comparative analysis. The corresponding quantitative results are summarized in Table \uppercase\expandafter{\romannumeral7}.  Compared to the standard MPD-PPO in Fig. 5, Fig. 8 displays the MPD-PPO without Differentiated Clipping results.

Experimental results confirm that although removing the differentiated clipping leads to significant performance degradation in film width optimization, the overall performance of our proposed method remains superior to the original PPO algorithm. Furthermore, since the MPD-PPO algorithm incorporates independent advantage estimation for each branch network, it demonstrates better optimization effectiveness than the PPO algorithm with branched networks from previous experiments, even without the differentiated clipping mechanism.
\begin{table}[htbp]
\centering
\caption{Optimization Results Of Removing Differentiated Clipping. }
\label{tab:three_line}
\begin{tabularx}{0.48\textwidth}{>{\raggedright\arraybackslash}X c}
\toprule
\textbf{ Algorithm}  & \textbf{Average Optimize Step} \\
\midrule
MPD-PPO  & 22 \\
MPD-PPO Without Differentiated Clipping  & 87 \\
PPO  & 100 (Failed) \\
\bottomrule
\end{tabularx}
\end{table} 

  \begin{figure}[h]
    \centering
    \includegraphics[width=0.98\linewidth]{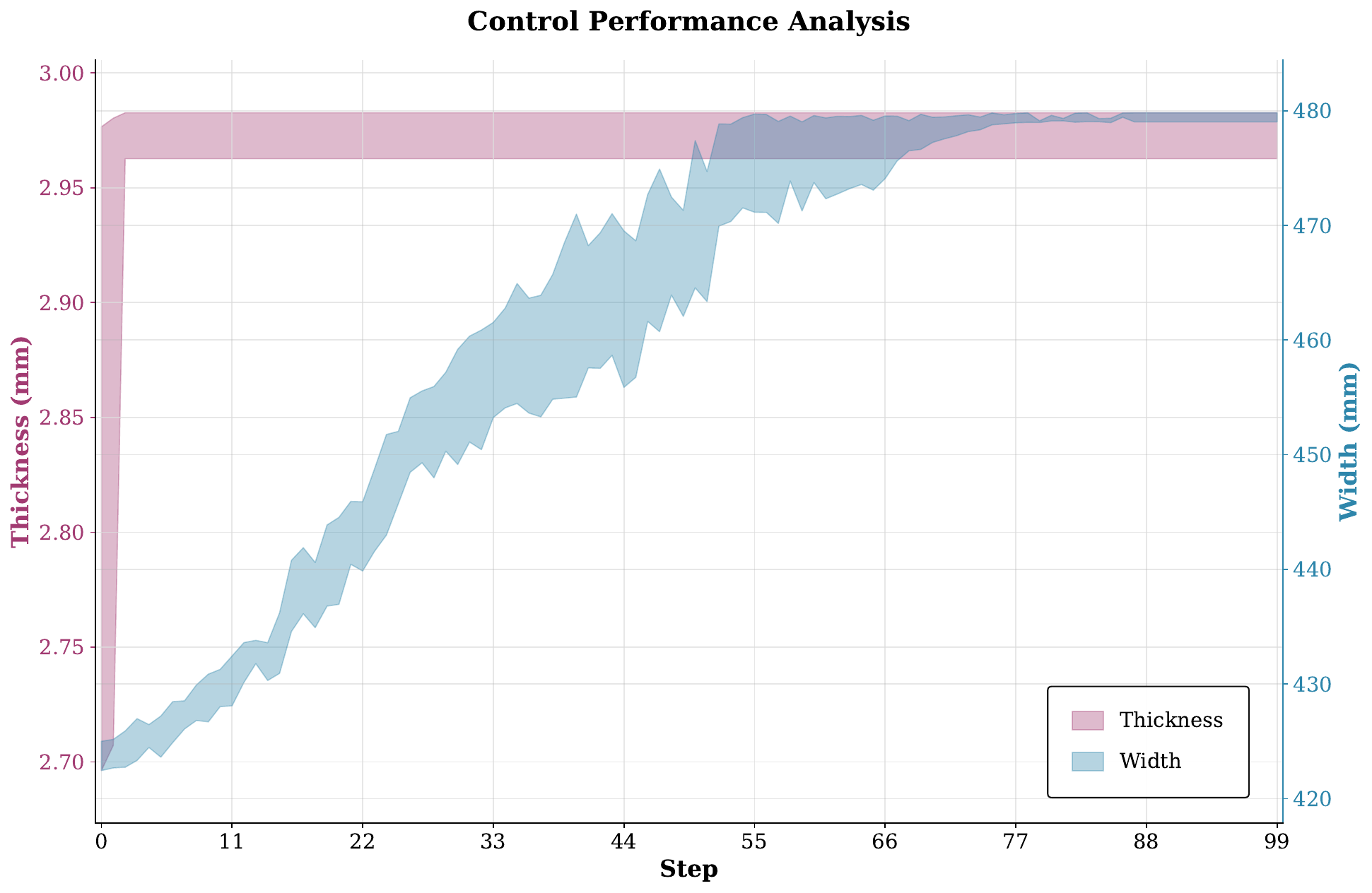}
    \caption{Optimization Results of MPD-PPO with Differentiated Clipping Removed. }
    \label{fig:placeholder}
\end{figure}

\subsubsection{Impact of Reward-Function Modules on Optimization Performance}

To isolate the contribution of individual reward components, we performed a comparative analysis using width control as the test case, chosen due to the less pronounced contrast in thickness scale. We derived all evaluated reward functions by systematically removing components from our proposed comprehensive function: Reward 1 uses only the Error Reward; Reward 2 combines Error and Progress Rewards; Reward 3 excludes the Steady-State Reward; Reward 4 is the complete function, integrating an action smoothness penalty with error and progress rewards to balance control accuracy and operational stability. The corresponding results are presented in Table \uppercase\expandafter{\romannumeral8} and Fig. 9.
\begin{table}[htbp]
\centering
\caption{The Impact of Reward-Function Modules on Optimization Results. }
\label{tab:three_line}
\begin{tabularx}{0.48\textwidth}{>{\raggedright\arraybackslash}X c}
\toprule
\textbf{ Algorithm}  & \textbf{Average Optimize Step} \\
\midrule
Reward 1 (Use Only The Error Reward)   & 74 \\
Reward 2 (Combines Error And Progress Reward) & 100 (Falied)  \\
Reward 3 (Excludes The Steady-State Reward) & 22 \\
Reward 4 (The Complete Reward Function) & 16 \\
\bottomrule
\end{tabularx}
\end{table} 
  \begin{figure}
    \centering
    \includegraphics[width=0.98\linewidth]{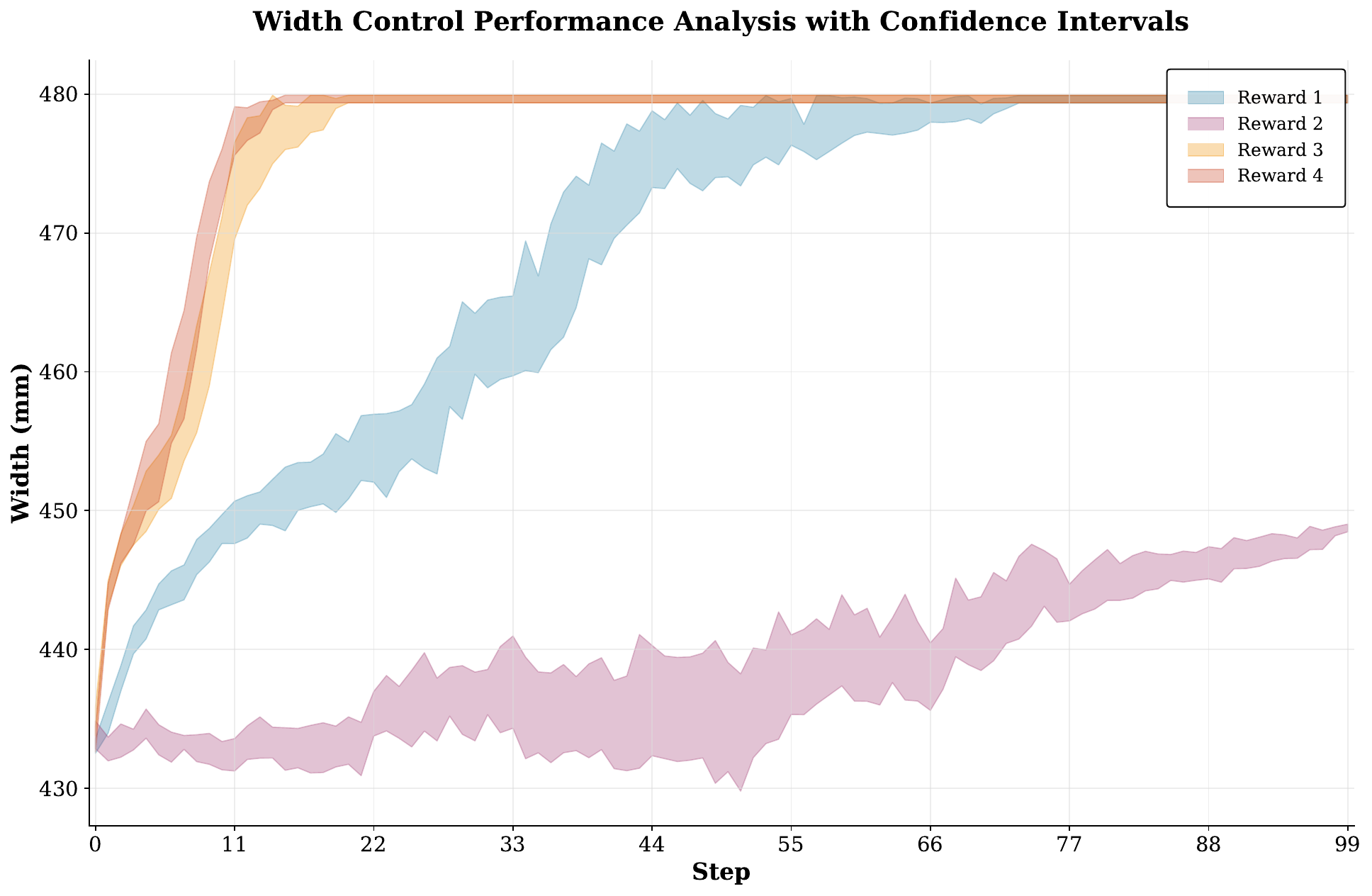}
    \caption{The Impact of Reward-Function Modules on Optimization Results. }
    \label{fig:placeholder}
\end{figure}
The experimental results confirm that every component of the reward function contributes to the optimization of the width parameters, demonstrating the synergy of combining all the individual modules.

\subsection{Experimental Validation on a Real-World Rubber Tyre Production Line}

To rigorously assess the algorithm's effectiveness under real-world industrial conditions, this study conducted testing in collaboration with  MESNAC,  on one of its operational production lines, the same industrial equipment as illustrated in Fig. 1. Process data were collected in real-time via PLC communication at a  frequency of 0.2 \emph{ms} intervals, directly from sensors and control systems. This \emph{in-situ}  validation strategy guarantees a high degree of fidelity to the actual production environments, thereby providing dependable evidence for assessing the algorithm's performance. The experimental results are presented in Fig. 10 and Fig. 11. Experimental results from the production line demonstrate the robustness of the agent's control strategy. In response to both positive and negative setpoint adjustments in width, the agent generated appropriate control actions that effectively guided the system state toward the target. The regulation process exhibited no significant overshoot, indicating favorable convergence characteristics.

  \begin{figure}[h]
    \centering
    \includegraphics[width=0.98\linewidth]{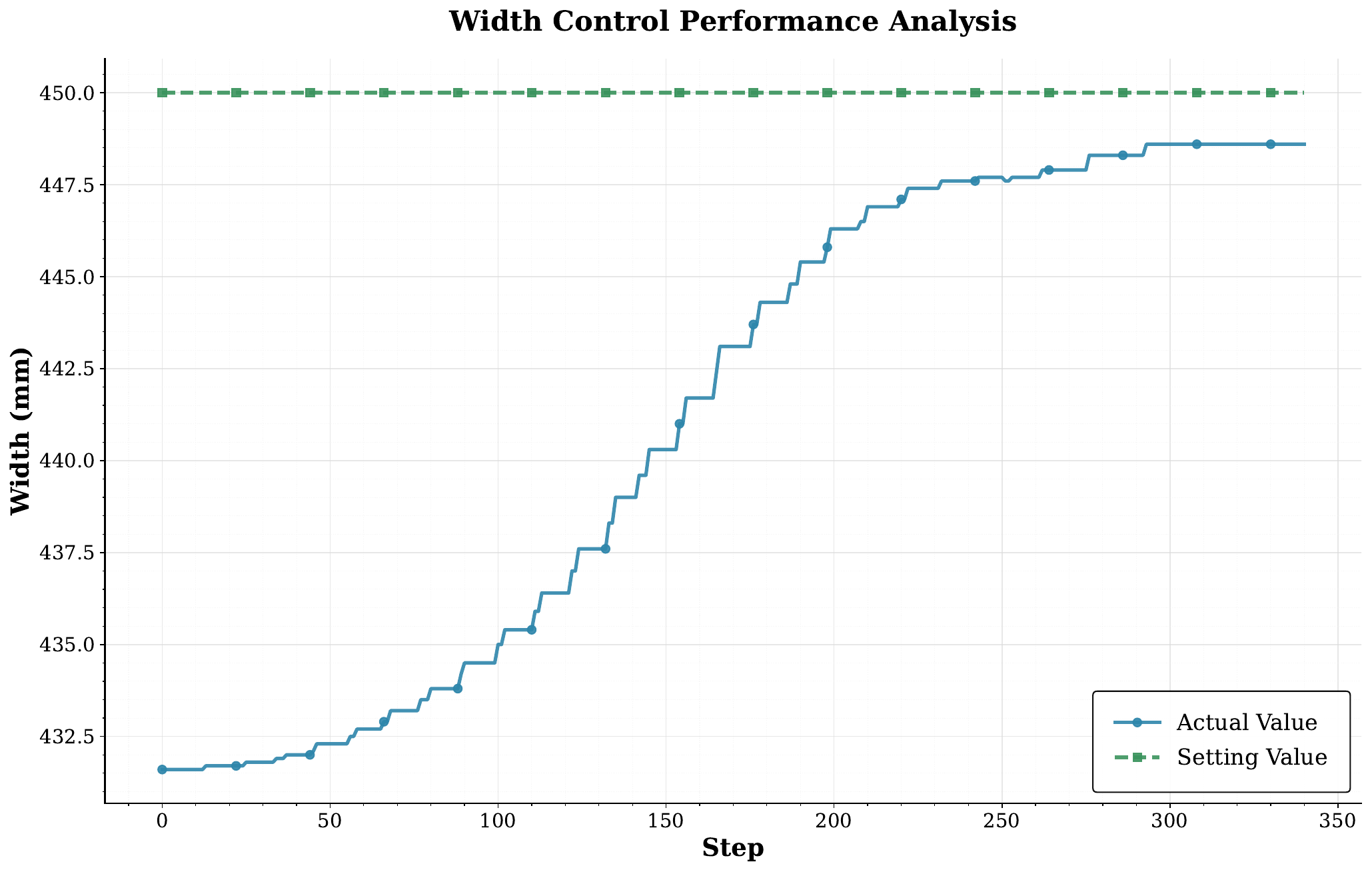}
    \caption{Validation Results From A Real Rubber Tyre Production Line 1. }
    \label{fig:placeholder}
\end{figure}
  \begin{figure}[h] 
    \centering
    \includegraphics[width=0.98\linewidth]{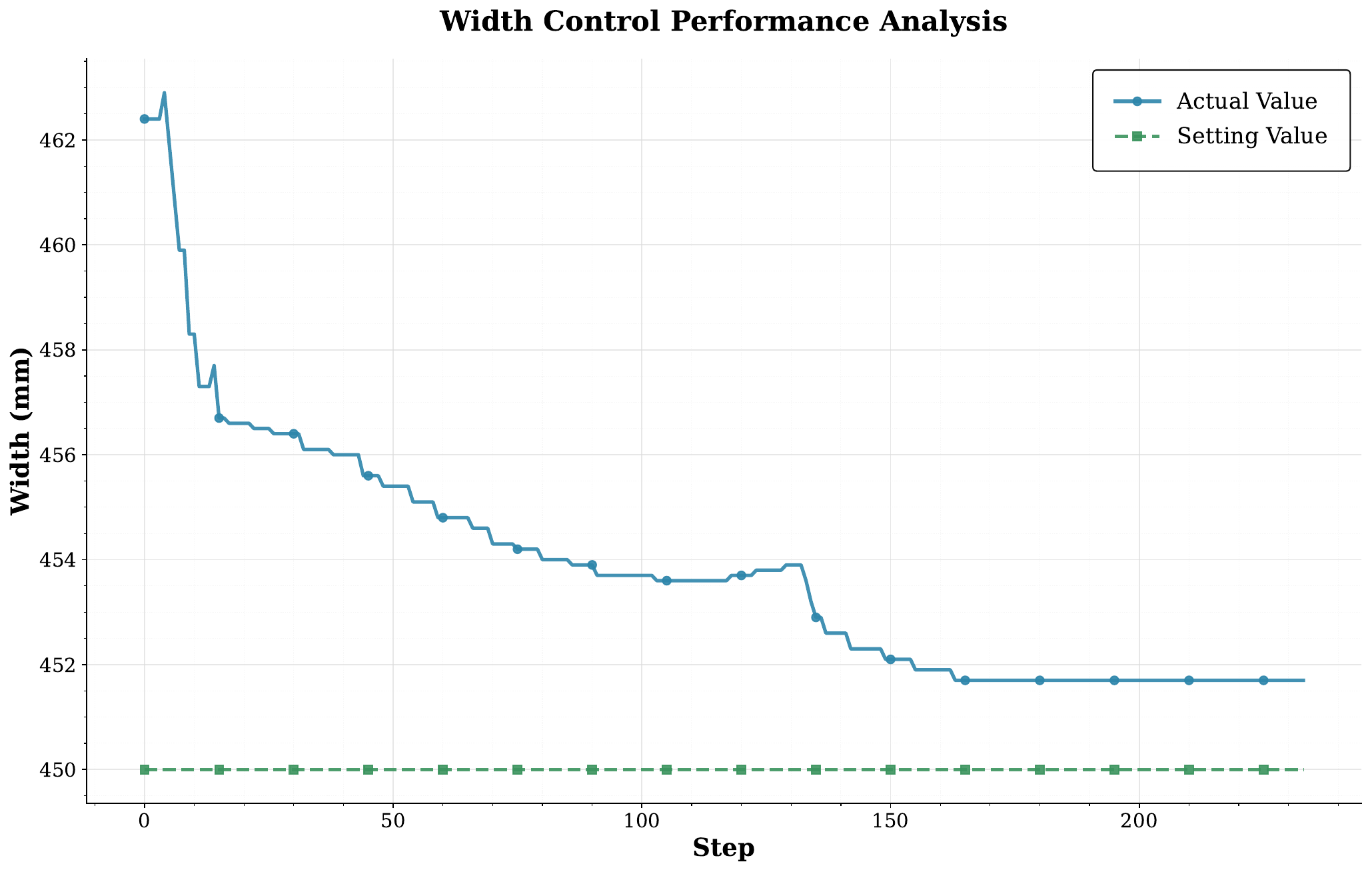}
    \caption{Validation Results From A Real Rubber Tyre Production Line  2. }
    \label{fig:placeholder}
\end{figure}

\section{Conclusion}

In this research, we propose a novel intelligent control framework that seamlessly integrates deep reinforcement learning with predictive modeling to address multi-objective parameter optimization in complex industrial processes. 

The proposed solution combines the predictive capabilities of LSTNet, which effectively captures both short-term fluctuations and long-term trends in complex time series data, with the Multi-path Differentiated Clipping Proximal Policy Optimization (MPD-PPO) algorithm. MPD-PPO employs a distributed policy architecture and an adaptive clipping mechanism to achieve differentiated and precise control. Furthermore, a composite reward function was designed to balance the competing objectives of accuracy, efficiency, and stability. Collectively, this research provides a scalable and adaptive solution for industrial autonomous optimization, demonstrating enhanced real-time performance and effective handling of multi-objective trade-offs and dynamic state processing. The methodology thus offers a promising pathway for advancing reinforcement learning applications in industrial informatics and control systems.

% References
\printbibliography

%\vspace{-1cm}
% \begin{IEEEbiography}[{\includegraphics[width=1in,height=1.25in,clip,keepaspectratio]{photo-men.eps}}]
% {First A. Author1} and the other authors may include biographies at the end of regular papers. The first paragraph may contain a place and/or date of birth (list place, then date). Next, the author's educational background is listed. The degrees should be listed with type of degree in what field, which institution, city, state or country, and year degree was earned. The author's major field of study should be lower-cased.

% The second paragraph uses the pronoun of the person (he or she) and not the author's last name. It lists military and work experience, including summer and fellowship jobs. Job titles are capitalized. The current job must have a location; previous positions may be listed without one. Information concerning previous publications may be included.

% The third paragraph begins with the author's title and last name (e.g., Dr. Smith, Prof. Jones, Mr. Kajor, Ms. Hunter). List any memberships in professional societies other than the IEEE. Finally, list any awards and work for IEEE committees and publications. If a photograph is provided, the biography will be indented around it. The photograph is placed at the top left of the biography. Personal hobbies will be deleted from the biography.
% \end{IEEEbiography}

\end{document}